\documentclass[runningheads]{llncs}
\usepackage[T1]{fontenc}
\usepackage{graphicx}
\usepackage{microtype}
\usepackage{graphicx}
\usepackage{subcaption}
\usepackage{booktabs} % for professional tables
\usepackage{multirow}    % \multirow
\usepackage{array}       % better column formatting
\usepackage[table]{xcolor}
\usepackage{tabularx}

\usepackage{amsmath}
\usepackage{amssymb}
\usepackage{mathtools}
\usepackage{makecell}
\usepackage{cite}
\begin{document}
%
% \title{Contribution Title\thanks{Supported by organization x.}}
\title{Cross-Dataset Transfer and Unknown-Class Detection in Imbalanced SAR Ship Classification}

%
%\titlerunning{Abbreviated paper title}
% If the paper title is too long for the running head, you can set
% an abbreviated paper title here
%
\author{Ch Muhammad Awais\inst{1}\orcidID{0009-0001-4589-4103} \and
Marco Reggiannini\inst{1}\orcidID{0000-0002-4872-9541} \and
Davide Moroni\inst{1}\orcidID{0000-0002-5175-5126} \and
Giulio Del Corso\inst{1}\orcidID{0000-0003-4604-2006}}
\authorrunning{C. Awais et al.}
% First names are abbreviated in the running head.
% If there are more than two authors, 'et al.' is used.
%
% \institute{Princeton University, Princeton NJ 08544, USA \and
% Springer Heidelberg, Tiergartenstr. 17, 69121 Heidelberg, Germany
% \email{lncs@springer.com}\\
% \url{http://www.springer.com/gp/computer-science/lncs} \and
% ABC Institute, Rupert-Karls-University Heidelberg, Heidelberg, Germany\\
% \email{\{abc,lncs\}@uni-heidelberg.de}}
\institute{Institute of Science and Technology, National Research Council, Pisa, Italy
\email{fname.lname@isti.cnr.it}}
\maketitle              % typeset the header of the contribution

\begin{abstract}
% The abstract should briefly summarize the contents of the paper in
% 150--250 words.
Ship classification from Synthetic Aperture Radar (SAR) imagery is a critical computer vision task, yet the robustness of models under deployment shifts remains unclear. While models are often trained on one dataset and deployed on another, we lack a comprehensive understanding of their cross-dataset generalization. To address this, we evaluate six pretrained models on two SAR ship datasets in three settings: in-domain classification, cross-dataset transfer, and unknown-class detection. For unknown detection, we hold out all classes one at a time. In-domain, SARDet100K gives the best balanced accuracy on both datasets (73.4\% on FUSARShip and 53.2\% on OpenSARShip). In cross-dataset transfer, we observe strong failures: some models show moderate accuracy but near-chance balanced accuracy (for example, 64.5\% accuracy but 33.3\% balanced accuracy for OpenSARShip to FUSARShip). In unknown detection, performance depends on the held-out class and dataset, while MC-dropout variance is often close to random. These findings show that cross-dataset generalization in SAR remains limited and that task-specific uncertainty scores are often more informative than MC-dropout variance for held-out-class detection, although their relative ranking depends on the dataset and held-out class.

\end{abstract}
\section{Introduction}
Synthetic aperture radar (SAR) is a core sensing technology for Earth Observation (EO) because it operates in day and night conditions and can penetrate clouds, haze, and rain. This makes SAR especially important for maritime applications where optical imagery is often unreliable. SAR is used in ship traffic monitoring \cite{li2022deep, li2023comprehensive}, illegal fishing surveillance \cite{10242360}, border and coast guard operations \cite{topouzelis2008oil, zilman2004speed}, search-and-rescue support, and early warning systems for maritime risk \cite{jmse7070202}. In these applications, reliable ship-type classification is not only a benchmark problem but also an operational requirement \cite{awais2025survey}. Classification errors can delay response actions or trigger false alarms in safety-critical workflows. For this reason, models must be evaluated not only for accuracy but also for reliability under changing operational conditions \cite{10701968}.

% \begin{figure}[t]
%     \centering
%     \includegraphics[width=0.5\linewidth]{imta_icpr_graphical_v4.pdf}
%     \caption*{Exploring uncertainty in SAR Ship Classification}
%     % \caption*{Unknown-class detection setup: training on known ship types while evaluating the detection of a held-out class at test time.}
%     \label{fig:teaser}
% \end{figure}

The methodology for SAR ship classification has progressed through clear stages. Early work relied on classical machine learning with hand-crafted features \cite{10446741}. This was followed by deep learning models trained end-to-end for specific datasets. More recently, transfer learning with ImageNet-pretrained backbones became common with limited SAR-labeled data \cite{awais2025survey}. The current shift is toward domain-specific foundation models, which are trained at larger scale and are expected to provide better transferable representations \cite{xiao2025foundation, yuan2021tokens}. However, these models are still new in SAR, and their downstream behavior under realistic deployment shift is not yet well understood \cite{awaisfusion}.

This gap motivates our study. We benchmark EO foundation-model \cite{lu2025vision} representations on SAR ship classification and evaluate not only predictive performance but also epistemic uncertainty under shift. This is important because models are often trained on one dataset and deployed on another with different sensors, acquisition geometry, preprocessing pipelines, and class balance. Under such changes, a model can retain moderate accuracy while failing on minority classes or unknown targets \cite{del2025shedding}. We therefore ask: how well do foundation-model features for SAR ship classification perform in-domain, across datasets, and when unknown classes appear?

To study this question, we introduce the SAR-Shift-Open Benchmark (SSOB), a comprehensive evaluation framework for six pretrained EO foundation models: DOFA, Prithvi, ScaleMAE, SSL4EO, SARDet100K, and SARJEPA. SSOB covers three settings: in-domain classification (ID), cross-dataset transfer (CD), and unknown-class detection (UC). In the unknown-class setting, we hold out Cargo, Fishing, and Tanker one at a time. We compare four uncertainty signals selected among the non-intrusive strategies that can be applied without systematic model modifications: \emph{softmax confidence} \cite{hendrycks2016baseline} (commonly used even if often positively biased), \emph{MC-dropout variance} \cite{gal2016dropout} (easy extension at the cost of multiple forward passes), \emph{class-oriented uncertainty}, and \emph{domain-aware uncertainty} \cite{ovadia2019can,geng2020recent} (more focused on anomaly detection).

%We compare four uncertainty signals: softmax confidence \cite{hendrycks2016baseline}, MC-dropout variance \cite{gal2016dropout}, class-oriented uncertainty, and domain-aware uncertainty \cite{ovadia2019can,geng2020recent}.

% To answer this question, we introduce \textbf{SAR-Shift-Open Benchmark} (SSOB), a unified protocol for six pretrained EO foundation models (DOFA, Prithvi, ScaleMAE, SSL4EO, SARDet100K, SARJEPA) in three settings: (i) in-domain (ID), (ii) cross-dataset (CD) transfer, and (iii) unknown-class (UC) detection, we hold out Cargo, Fishing, and Tanker one at a time for UC. We compare four uncertainty signals: softmax confidence \cite{hendrycks2016baseline}, MC-dropout variance \cite{gal2016dropout}, and two task-specific scores, class-oriented uncertainty ($s_{\text{class}}$), and domain-aware uncertainty ($s_{\text{domain}}$), motivated by the need for reliable uncertainty under distribution shift and open-set conditions \cite{ovadia2019can,geng2020recent}.

\paragraph{Contributions.}
% \begin{enumerate}
%     \item \textbf{Downstream evaluation of SAR foundation models.}
%     We provide a controlled benchmark for testing foundation-model representations on a real SAR downstream task, with repeated-run reporting.

%     \item \textbf{Quantified cross-dataset failure mode.}
%     We show that cross-dataset transfer can degenerate to majority-class prediction. For OpenSARShip to FUSARShip, Prithvi reaches $0.645 \pm 0.003$ accuracy but only $0.333 \pm 0.002$ balanced accuracy.

%     \item \textbf{Epistemic uncertainty benchmarking under unknown classes.}
%     We evaluate unknown detection with three held-out classes and show that uncertainty quality depends strongly on held-out class and dataset.

%     \item \textbf{Model ranking under robust metrics.}
%     SARDet100K provides the strongest in-domain balanced accuracy on both FUSARShip ($0.734 \pm 0.040$) and OpenSARShip ($0.532 \pm 0.034$).
% \end{enumerate}
\begin{enumerate}
    \item We introduce SSOB, a comprehensive framework for evaluating pretrained representations on SAR ship classification under in-domain, cross-dataset, and held-out-class unknown settings.
    \item We show that cross-dataset evaluation can hide failure when reported with accuracy alone, which motivates balanced metrics and per-class analysis.
    \item We compare four post-hoc uncertainty scores for held-out-class detection and show that their behavior is strongly class and dataset dependent.
\end{enumerate}
We do not propose a new backbone or training method. Our goal is to evaluate pretrained representations under a shared framework across matched, shifted, and held-out-class settings. 

%old
%Section~\ref{sec:method} describes SSOB and the uncertainty protocol.
%Section~\ref{sec:results} reports ID, CD, and UC results.
%Section~\ref{sec:discussion} discusses implications and limitations.
%Section~\ref{sec:conclusion} concludes.
%NEW
The paper is structured as follows: Section~\ref{sec:method} introduces the SSOB evaluation framework and our selected uncertainty metrics. Section~\ref{sec:results} details the experimental results across all three distribution shift settings, followed by a discussion of operational implications and limitations in Section~\ref{sec:discussion}. Section~\ref{sec:conclusion} concludes the work.

\section{Methodology}
\label{sec:method}

\begin{figure}[htbp]
    \centering
    \includegraphics[width=1\linewidth]{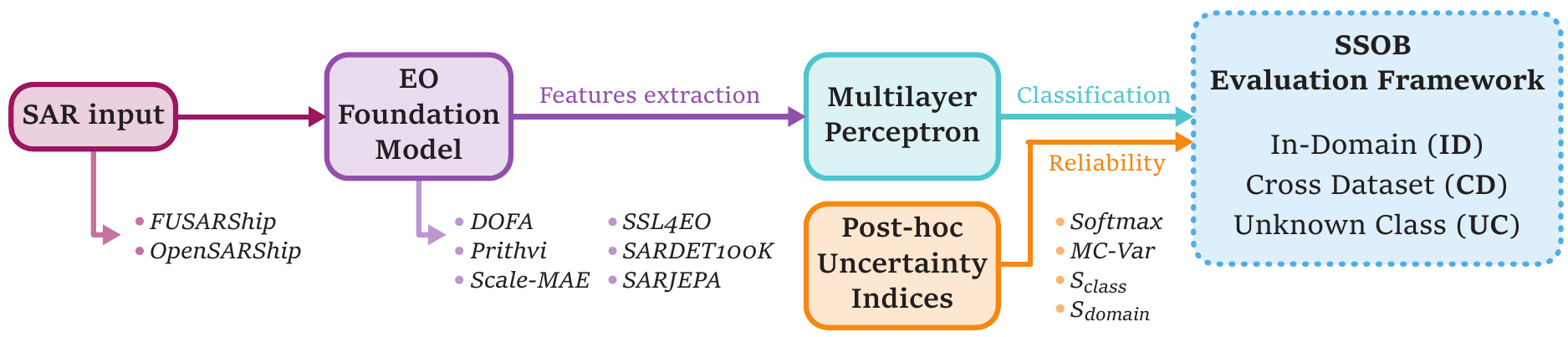}
    \caption{Overview of the evaluation pipeline used in this study. SAR images are encoded with an Earth Observation (EO) foundation model to extract representations, followed by training of a shared classifier head. The trained model is then evaluated in three settings: in-domain, cross-dataset, and unknown-class settings. In-domain and cross-dataset branches report classification performance, while the unknown-class branch is used for uncertainty-based analysis.}
    \label{fig:methodology}
\end{figure}

% \vspace{-0.2cm}
Figure~\ref{fig:methodology} summarizes the study pipeline. We evaluate SAR foundation-model representations by freezing each backbone and training a lightweight classifier head under a fixed setting.

\subsection{Foundation Models}
We benchmark six EO foundation models, DOFA~\cite{xiong2024neural}, Prithvi~\cite{jakubik2023foundation}, ScaleMAE~\cite{reed2023scale}, SSL4EO~\cite{wang2023ssl4eo}, SARDet100K~\cite{li2024sardet}, and SARJEPA~\cite{li2024predicting}, on SAR ship classification under a fixed training and evaluation setting. DOFA is a multimodal EO model designed to handle varying sensor types and channel configurations; Prithvi is a transformer-based geospatial model pretrained on HLS imagery; ScaleMAE learns scale-aware remote-sensing features via masked autoencoding; SSL4EO provides self-supervised weights trained on the SSL4EO-S12 Sentinel-1/Sentinel-2 corpus; SARDet100K is pretrained on large-scale SAR object detection; and SARJEPA is a SAR self-supervised model based on joint-embedding predictive learning.

\subsection{Datasets}
\begin{table}[ht]
    \centering
    \caption{Physical properties and instance counts; FUSARShip (FShip); OpenSARShip (OShip). 'Res.' = spatial resolution.}
    \label{tab:data_information}
    \begin{tabular}{lcccc}
       \hline
       \textbf{Dataset} & \textbf{Res.} & \textbf{Cargo} & \textbf{Tanker} & \textbf{Fishing} \\
       \hline
       FShip & $\sim$1.5 m & 1693 & 785 & 148 \\
       OShip & $\sim$20 m & 5303 & 139 & 1825 \\
       \hline
    \end{tabular}
\end{table}
We evaluate all models on FUSARShip~\cite{hou2020fusar} and OpenSARShip~\cite{huang2017opensarship}, two public SAR ship datasets with different acquisition characteristics. FUSARShip is built from high-resolution GF-3 SAR images, while OpenSARShip is based on Sentinel-1 ship images. To ensure comparability and avoid extreme imbalance, we restrict both datasets to their shared ship categories (Cargo, Tanker, Fishing) and apply a consistent class mapping (Table~\ref{tab:data_information}). In all experiments, the backbone is frozen and only a lightweight classifier head is trained (described in the methodology section), so performance differences primarily reflect representation quality rather than end-to-end fine-tuning.

\subsection{Classification Pipeline}
Let a sample be $(x,y)$, where $x$ is a SAR image and $y \in \{\text{Cargo}, \text{Tanker}, \text{Fishing}\}$. For a pretrained model $m$, the backbone $\phi_m(\cdot)$ produces a feature vector $z=\phi_m(x)$, and a classifier head $h_\theta(\cdot)$ maps $z$ to logits. The classifier is a multilayer perceptron consisting of a 512-dimensional hidden layer with ReLU nonlinearity, dropout probability 0.2 and a final linear projection to the class logits. The backbone is frozen all the time; only the classifier head is optimized.

Training uses cross-entropy loss, the Adam optimizer, learning rate $10^{-3}$, weight decay $10^{-4}$, and batch size 512 over up to 100 epochs. Early stopping monitors validation loss with patience 20 (minimum improvement $10^{-4}$). Each configuration is evaluated over 5 data splits and 5 random seeds. %The same repeated-run protocol applies across in-domain, cross-dataset, and unknown-class evaluation settings.

\subsection{Uncertainty Scores}
\label{sec:uncertainty_scores}
Let $x$ denote an input SAR image. The frozen backbone $\phi_m$ maps $x$ to a feature vector $z=\phi_m(x)\in\mathbb{R}^d$ with d ranging from 384 for SSL4EO to 2048 of SARDet100K backbones. Feature vectors are precomputed and cached during inference. A linear classifier head produces logits $a=Wz$ and class probabilities $p=\mathrm{softmax}(a)$.

We evaluate four post-hoc uncertainty quantification methods:

\paragraph{Maximum softmax probability (Softmax).}
We use the complement of maximum softmax probability as a baseline uncertainty signal:
\[
u_{\text{soft}}(x)=1-\max_k p_k(x).
\]
% This score reflects predictive confidence; higher values indicate lower model confidence but is known to be positively biased and tends to produce overconfident reliability predictions \cite{del2025shedding}.
This score reflects predictive confidence; higher values indicate lower model confidence, but it is known to be positively biased and tends to produce overconfident reliability predictions \cite{del2025shedding}.

\paragraph{Classification Head MC-dropout variance (MC-Var)}
Monte Carlo dropout estimates epistemic (model) uncertainty by performing stochastic inference. With dropout enabled during inference, we execute $T=10$ stochastic forward passes through the frozen backbone and classifier to make a fair comparison, producing predictions $\{p^{(t)}(x)\}_{t=1}^T$. The choice of using 10 forward passes is justified by \cite{kendall2017uncertainties}; while this small number of simulations may not guarantee perfect stability, it is sufficient to identify the most unreliable predictions. We compute class-wise variance and average across classes:
\[
u_{\text{mc}}(x)=\frac{1}{K}\sum_{k=1}^K \mathrm{Var}_{t}\!\left(p^{(t)}_k(x)\right),
\]
where $K=3$ is the number of classes. Higher values indicate greater prediction variability across stochastic forward passes.

\paragraph{Class-oriented uncertainty ($s_{\text{class}}$).}
We compute class prototypes from the source training set as the mean feature representation per class:
\[
\mu_c=\frac{1}{|D_c|}\sum_{(x_i,y_i)\in D_c}\phi_m(x_i), \qquad c\in\{1,2,3\},
\]
where $D_c$ is the subset of training samples with label $c$. Class-oriented uncertainty is defined as the minimum Euclidean distance to any class prototype:
\[
s_{\text{class}}(x)=\min_{c\in\{1,2,3\}} \left\lVert z-\mu_c\right\rVert_2.
\]
% This score increases for samples that lie distant from all learned class clusters in feature space, indicating atypicality relative to known classes. 
%%READ%%
% Prototype-based scores are more reliable because they directly measure distance from the known training distribution in feature space, rather than deriving uncertainty from a classifier boundary that was never trained to represent out-of-distribution inputs. In addition, has been proved \cite{jiang2018trust} that they tend to converge to the optimal Bayesian approximator even without further re-training.
This score increases for samples that lie distant from all learned class clusters in feature space, indicating atypicality relative to known classes. Prototype-based scores are more reliable because they directly measure distance from the known training distribution in feature space, rather than deriving uncertainty from a classifier boundary that was never trained to represent out-of-distribution inputs. In addition, it has been proven \cite{jiang2018trust} that they tend to converge to the optimal Bayesian approximator even without further re-training.

\paragraph{Domain-aware uncertainty ($s_{\text{domain}}$).}
From the same source training features, we compute a global domain center as the mean across all training samples:
\[
\mu_{\text{dom}}=\frac{1}{|D_{\text{train}}|}\sum_{(x_i,y_i)\in D_{\text{train}}}\phi_m(x_i).
\]
Domain-aware uncertainty is the Euclidean distance to this center:
\[
s_{\text{domain}}(x)=\left\lVert z-\mu_{\text{dom}}\right\rVert_2.
\]
This score increases as test samples deviate from the training-domain feature distribution, capturing domain shift. %All four scores are normalized so that larger values indicate higher uncertainty.

\subsection{Evaluation Framework}
\label{sec:eval_protocol}
We evaluate model performance across three distinct settings: in-domain (ID), cross-dataset (CD), and unknown-class (UC).

\subsubsection{In-Domain Evaluation}
Models are trained and tested on the same dataset using five-fold cross-validation. This setting measures supervised performance under distribution match.

\subsubsection{Cross-Dataset Evaluation}
We evaluate transfer learning by training on one dataset and testing on the other, reporting both directions (FUSARShip to OpenSARShip and OpenSARShip to FUSARShip). This setting measures robustness to dataset shift while maintaining known-class structure.

\subsubsection{Unknown-Class Evaluation}
To simulate out-of-distribution detection, we conduct unknown-class experiments by holding out each class during training. For each class $c \in \{\text{Cargo, Tanker, Fishing}\}$, all samples with label $c$ are removed from the training set, and the classifier head is trained on the two remaining known classes. At test time, evaluation is performed on the full test set with class $c$ treated as unknown. Uncertainty scores are computed for all test samples, and unknown-detection performance is quantified via AUROC to measure the ability of each score to distinguish held-out samples from known classes. This procedure is repeated three times to provide independent results for each held-out class.

\section{Results}
\label{sec:results}
For classification, we report accuracy, balanced accuracy, macro-F1, and per-class recall. Balanced accuracy and macro-F1 serve as primary metrics under class imbalance.
% For uncertainty evaluation, we report AUROC (Area Under the Receiver Operating Characteristic Curve) as a threshold-free metric. In the unknown-class setting, AUROC quantifies the discriminative ability of each uncertainty score: a score of 0.5 (~50) indicates random performance, while 1.0 (~100) indicates perfect separation between known and unknown samples.
For uncertainty evaluation, we report the AUROC (Area Under the Receiver Operating Characteristic curve) of the uncertainty metrics against the agreement between model predictions and ground truth. This is a threshold-free metric bounded in [0,100\%], where the maximum value corresponds to an uncertainty score that perfectly identifies mismatched cases (i.e., model prediction $\ne$ ground truth), 50\% represents random performance (i.e., no discriminative ability), and values in [0,50\%] correspond to indices that tend to assign higher uncertainty to correct predictions rather than to misclassified ones.

\subsection{In-domain classification}
\begin{table}[htbp]
\centering
\definecolor{colhdr}{RGB}{220, 232, 248}
\renewcommand{\arraystretch}{1.25}
\caption{In-domain results. Best value per metric is \textbf{bold}.
Abbreviations: SD=SARDet100K, DF=DOFA, SJ=SARJEPA, S4=SSL4EO, SM=ScaleMAE, PV=Prithvi,
Acc=Accuracy, BAcc=Balanced Accuracy.}
\label{tab:id_results}
\begin{tabularx}{\linewidth}{l *{3}{>{\centering\arraybackslash}X} |*{3}{>{\centering\arraybackslash}X}}
\toprule
& \multicolumn{3}{c}{FUSARShip}
& \multicolumn{3}{c}{OpenSARShip} \\
\cmidrule(lr){2-4} \cmidrule(lr){5-7}
Model & Acc & BAcc & Macro-F1 & Acc & BAcc & Macro-F1 \\
\midrule
SD & \textbf{75.68$\pm$1.97} & \textbf{73.42$\pm$3.98} & \textbf{75.40$\pm$3.16} & \textbf{76.05$\pm$0.58} & \textbf{53.22$\pm$3.38} & \textbf{56.66$\pm$3.37} \\
DF & 69.58$\pm$1.63 & 56.57$\pm$2.68 & 59.32$\pm$2.06 & 74.62$\pm$0.54 & 44.38$\pm$2.62 & 47.09$\pm$3.41 \\
SJ & 69.20$\pm$2.28 & 58.34$\pm$3.49 & 61.26$\pm$3.30 & 74.36$\pm$0.43 & 45.60$\pm$2.66 & 48.31$\pm$2.96 \\
S4 & 68.94$\pm$1.65 & 57.51$\pm$2.33 & 60.38$\pm$2.41 & 73.71$\pm$0.49 & 38.17$\pm$1.51 & 37.62$\pm$2.62 \\
SM & 66.49$\pm$1.66 & 50.91$\pm$4.75 & 53.53$\pm$5.12 & 73.16$\pm$0.21 & 38.69$\pm$2.09 & 36.99$\pm$3.17 \\
PV & 66.33$\pm$1.81 & 47.21$\pm$3.93 & 49.58$\pm$5.03 & 73.12$\pm$0.06 & 35.95$\pm$1.01 & 32.92$\pm$1.76 \\
\bottomrule
\end{tabularx}
\end{table}
% \begin{table}[ht]
% \centering
% \caption{In-domain results. Best value for each metric is bold. 
% Abbreviations: SD=SARDet100K, DF=DOFA, SJ=SARJEPA, S4=SSL4EO, SM=ScaleMAE, PV=Prithvi, Acc=Accuracy, BAcc=Balanced Accuracy.}
% \label{tab:id_results}
% % \setlength{\tabcolsep}{4pt}
% \begin{tabular}{lccc}
% \toprule
% Model & Acc & BAcc & Macro-F1 \\
% \midrule
% \multicolumn{4}{c}{FUSARShip}\\
% SD & \textbf{75.68 $\pm$ 1.97} & \textbf{73.42 $\pm$ 3.98} & \textbf{75.40 $\pm$ 3.16} \\
% DF & 69.58 $\pm$ 1.63 & 56.57 $\pm$ 2.68 & 59.32 $\pm$ 2.06 \\
% SJ & 69.20 $\pm$ 2.28 & 58.34 $\pm$ 3.49 & 61.26 $\pm$ 3.30 \\
% S4 & 68.94 $\pm$ 1.65 & 57.51 $\pm$ 2.33 & 60.38 $\pm$ 2.41 \\
% SM & 66.49 $\pm$ 1.66 & 50.91 $\pm$ 4.75 & 53.53 $\pm$ 5.12 \\
% PV & 66.33 $\pm$ 1.81 & 47.21 $\pm$ 3.93 & 49.58 $\pm$ 5.03 \\
% \midrule
% \multicolumn{4}{c}{OpenSARShip}\\
% SD & \textbf{76.05 $\pm$ 0.58} & \textbf{53.22 $\pm$ 3.38} & \textbf{56.66 $\pm$ 3.37} \\
% DF & 74.62 $\pm$ 0.54 & 44.38 $\pm$ 2.62 & 47.09 $\pm$ 3.41 \\
% SJ & 74.36 $\pm$ 0.43 & 45.60 $\pm$ 2.66 & 48.31 $\pm$ 2.96 \\
% S4 & 73.71 $\pm$ 0.49 & 38.17 $\pm$ 1.51 & 37.62 $\pm$ 2.62 \\
% SM & 73.16 $\pm$ 0.21 & 38.69 $\pm$ 2.09 & 36.99 $\pm$ 3.17 \\
% PV & 73.12 $\pm$ 0.06 & 35.95 $\pm$ 1.01 & 32.92 $\pm$ 1.76 \\
% \bottomrule
% \end{tabular}
% \end{table}
In-domain results are shown in Table~\ref{tab:id_results}. SARDet100K is the top model on both datasets by balanced accuracy. 
%old
%The gap is large on FUSARShip (73.42 vs next best 58.34) and smaller but still clear on OpenSARShip (53.22 vs next best 45.60). 
%new 
The performance gap is substantial on FUSARShip (73.42\% compared to the second-highest, SARJEPA at 58.34\%) and remains distinct on OpenSARShip (53.22\% vs. 45.60\%).
%
%old
%This supports the benefit of SAR-specific pretraining for in-domain robustness.
%new
This supports the benefit of SAR-specific, task-aligned pretraining for in-domain robustness.

\subsection{Cross-dataset transfer}
% \begin{table}[htbp]
% \centering
% \caption{Cross-dataset results. Best value for each metric is bold.}
% \label{tab:cd_results}
% % \setlength{\tabcolsep}{4pt}
% \begin{tabular}{lccc}
% \toprule
% Model & Acc & BAcc & Macro-F1 \\
% \midrule
% \multicolumn{4}{c}{FUSARShip $\rightarrow$ OpenSARShip}\\
% SJ & \textbf{58.15$\pm$6.50} & 29.25$\pm$2.04 & 27.88$\pm$1.15 \\
% DF & 56.48$\pm$4.64 & \textbf{45.63$\pm$4.53} & 29.55$\pm$1.90 \\
% SM & 54.31$\pm$2.32 & 30.71$\pm$0.67 & 30.42$\pm$0.74 \\
% PV & 51.13$\pm$6.35 & 43.40$\pm$5.09 & 27.94$\pm$1.53 \\
% SD & 48.01$\pm$1.77 & 36.81$\pm$0.74 & \textbf{31.77$\pm$0.99} \\
% S4 & 29.42$\pm$2.03 & 39.71$\pm$5.70 & 21.19$\pm$1.55 \\
% \midrule
% \multicolumn{4}{c}{OpenSARShip $\rightarrow$ FUSARShip}\\
% PV & \textbf{64.49$\pm$0.32} & \textbf{33.34$\pm$0.16} & \textbf{26.30$\pm$0.36} \\
% SM & 64.11$\pm$0.77 & 33.10$\pm$0.40 & 26.06$\pm$0.19 \\
% S4 & 63.82$\pm$0.37 & 32.95$\pm$0.18 & 25.98$\pm$0.11 \\
% DF & 60.79$\pm$1.08 & 31.58$\pm$0.63 & 25.77$\pm$0.74 \\
% SJ & 57.11$\pm$2.95 & 30.46$\pm$1.25 & 25.18$\pm$0.77 \\
% SD & 53.63$\pm$3.98 & 31.98$\pm$4.07 & 25.97$\pm$2.24 \\
% \bottomrule
% \end{tabular}
% \end{table}
\begin{table}[ht]
\centering
\definecolor{colhdr}{RGB}{220, 232, 248}
\renewcommand{\arraystretch}{1.25}
\caption{Cross-dataset results. Best value per metric is \textbf{bold}.
Rows are ordered by F$\rightarrow$O rank; O$\rightarrow$F values are reordered to match.}
\label{tab:cd_results}
\begin{tabularx}{\linewidth}{l *{3}{>{\centering\arraybackslash}X} | *{3}{>{\centering\arraybackslash}X}}
\toprule
& \multicolumn{3}{c}{FUSARShip $\rightarrow$ OpenSARShip}
& \multicolumn{3}{c}{OpenSARShip $\rightarrow$ FUSARShip} \\
\cmidrule(lr){2-4} \cmidrule(lr){5-7}
Model & Acc & BAcc & Macro-F1 & Acc & BAcc & Macro-F1 \\
\midrule
SJ & \textbf{58.15$\pm$6.50} & 29.25$\pm$2.04 & 27.88$\pm$1.15 & 57.11$\pm$2.95 & 30.46$\pm$1.25 & 25.18$\pm$0.77 \\
DF & 56.48$\pm$4.64 & \textbf{45.63$\pm$4.53} & 29.55$\pm$1.90 & 60.79$\pm$1.08 & 31.58$\pm$0.63 & 25.77$\pm$0.74 \\
SM & 54.31$\pm$2.32 & 30.71$\pm$0.67 & 30.42$\pm$0.74 & 64.11$\pm$0.77 & 33.10$\pm$0.40 & 26.06$\pm$0.19 \\
PV & 51.13$\pm$6.35 & 43.40$\pm$5.09 & 27.94$\pm$1.53 & \textbf{64.49$\pm$0.32} & \textbf{33.34$\pm$0.16} & \textbf{26.30$\pm$0.36} \\
SD & 48.01$\pm$1.77 & 36.81$\pm$0.74 & \textbf{31.77$\pm$0.99} & 53.63$\pm$3.98 & 31.98$\pm$4.07 & 25.97$\pm$2.24 \\
S4 & 29.42$\pm$2.03 & 39.71$\pm$5.70 & 21.19$\pm$1.55 & 63.82$\pm$0.37 & 32.95$\pm$0.18 & 25.98$\pm$0.11 \\
\bottomrule
\end{tabularx}
\end{table}
Cross-dataset results in Table~\ref{tab:cd_results} show strong degradation from in-domain performance and clear direction asymmetry. In OpenSARShip to FUSARShip transfer, several models have moderate accuracy but near-chance balanced accuracy. The clearest example is Prithvi with 64.49 accuracy and 33.34 balanced accuracy. This is consistent with majority-class prediction and weak minority recall under shift. In FUSARShip to OpenSARShip transfer, the best balanced accuracy is higher (DOFA, 45.63) but remains far below in-domain values.

\subsection{Unknown-class detection}
% Table~\ref{tab:uc_summary} establishes a baseline using in-distribution balanced accuracy (BalAcc). Across all combinations, the SARDet100K model achieves the best known-only detection. BalAcc is consistently higher on FUSARShip, peaking at 0.9260 when Cargo is held out. OpenSARShip struggles comparatively, dropping to 0.6121 when Fishing is held out.
Table~\ref{tab:uc_summary} summarizes known-only classification performance (Acc, BalAcc, and Macro-F1) together with unknown-class detection AUROC. Across all combinations, the SARDet100K model achieves the strongest known-only classification. Performance is consistently higher on FUSARShip, peaking when Cargo is held out (97.17 Acc, 92.60 BalAcc, 94.53 Macro-F1), whereas OpenSARShip is more challenging, with the weakest known-only results when Fishing is held out (77.09 Acc, 61.21 BalAcc, 62.28 Macro-F1).
% \begin{table}[htbp]
% \centering
% \caption{Unknown-class AUROC. For each dataset and held-out class, each column reports the best model value for that score.}
% \label{tab:uc_summary}
% \small
% \setlength{\tabcolsep}{3pt} % Reduced from default 6pt
% \begin{tabular*}{\columnwidth}{@{\extracolsep{\fill}} l rrr cccc}
% \toprule
% Unknown & Acc (SD) & BAcc (SD) & M-F1 (SD) & Smax & MC-V & $s_{\text{cls}}$ & $s_{\text{dom}}$ \\
% \midrule
% \multicolumn{8}{l}{\textit{FUSARShip}} \\
% Cargo   & 97.2$\pm$0.9 & 92.6$\pm$1.6 & 94.5$\pm$1.6 & 62.9 (PV) & 51.7 (PV) & 59.6 (SD) & 55.4 (SD) \\
% Fishing & 97.7$\pm$0.3 & 88.7$\pm$3.0 & 91.6$\pm$1.4 & 42.1 (SD) & 49.0 (SD) & 66.3 (SJ) & 58.1 (SM) \\
% Tanker  & 75.9$\pm$1.7 & 70.5$\pm$2.7 & 71.0$\pm$2.5 & 42.5 (DF) & 50.2 (DF) & \textbf{87.9} (PV) & \textbf{86.2} (PV) \\
% \midrule
% \multicolumn{8}{l}{\textit{OpenSARShip}} \\
% Cargo   & 95.2$\pm$0.7 & 74.3$\pm$4.6 & 78.4$\pm$4.0 & 45.8 (SJ) & 49.8 (SJ) & 57.9 (DF) & 55.6 (DF) \\
% Fishing & 77.1$\pm$0.8 & 61.2$\pm$2.2 & 62.3$\pm$2.8 & 75.0 (PV) & 52.0 (SJ) & \textbf{82.9} (SM) & \textbf{83.7} (SM) \\
% Tanker  & 92.0$\pm$0.3 & 67.8$\pm$5.9 & 72.9$\pm$6.0 & 58.4 (SD) & 51.5 (SD) & 48.0 (SM) & 48.7 (SM) \\
% \bottomrule
% \end{tabular*}
% \end{table}
\begin{table}[htbp]
\centering
\footnotesize
\setlength{\tabcolsep}{2pt}
\renewcommand{\arraystretch}{1.3}
\caption{Unknown-class AUROC. For each dataset and held-out class, each column reports the best model value for that score.}
\label{tab:uc_summary}
\begin{tabular}{lccccccc}
\toprule
Unknown
  & \makecell{Acc \\ (SD)}
  & \makecell{BAcc \\ (SD)}
  & \makecell{Mac-F1 \\ (SD)}
  & Softmax & MC-Var
  & $s_{\text{class}}$ & $s_{\text{domain}}$ \\
\midrule
\multicolumn{8}{l}{\textit{FUSARShip}} \\[2pt]
Cargo
  & \makecell{97.17 \\ $\pm$0.92}
  & \makecell{92.60 \\ $\pm$1.63}
  & \makecell{94.53 \\ $\pm$1.64}
  & 62.87 (PV) & 51.72 (PV) & 59.58 (SD) & 55.35 (SD) \\
Fishing
  & \makecell{97.68 \\ $\pm$0.33}
  & \makecell{88.65 \\ $\pm$3.03}
  & \makecell{91.59 \\ $\pm$1.40}
  & 42.06 (SD) & 49.02 (SD) & 66.29 (SJ) & 58.05 (SM) \\
Tanker
  & \textit{\makecell{75.87 \\ $\pm$1.72}}
  & \textit{\makecell{70.49 \\ $\pm$2.73}}
  & \textit{\makecell{71.03 \\ $\pm$2.54}}
  & 42.51 (DF) & 50.17 (DF) & \textbf{87.94} (PV) & \textbf{86.19} (PV) \\
\midrule
\multicolumn{8}{l}{\textit{OpenSARShip}} \\[2pt]
Cargo
  & \makecell{95.21 \\ $\pm$0.66}
  & \makecell{74.27 \\ $\pm$4.56}
  & \makecell{78.38 \\ $\pm$3.96}
  & 45.83 (SJ) & 49.76 (SJ) & 57.91 (DF) & 55.55 (DF) \\
Fishing
  & \textit{\makecell{77.09 \\ $\pm$0.78}}
  & \textit{\makecell{61.21 \\ $\pm$2.16}}
  & \textit{\makecell{62.28 \\ $\pm$2.83}}
  & 74.98 (PV) & 51.96 (SJ) & \textbf{82.88} (SM) & \textbf{83.67} (SM) \\
Tanker
  & \makecell{91.98 \\ $\pm$0.28}
  & \makecell{67.81 \\ $\pm$5.94}
  & \makecell{72.86 \\ $\pm$6.04}
  & 58.36 (SD) & 51.50 (SD) & 48.00 (SM) & 48.67 (SM) \\
\bottomrule
\end{tabular}
\end{table}

% \begin{table}[htbp]
% \centering
% \caption{Unknown-class AUROC. For each dataset and held-out class, each column reports the best model value for that score.}
% \label{tab:uc_summary}
% \small
% % \setlength{\tabcolsep}{5pt}
% \begin{tabular}{lccccccc}
% \toprule
% Unknown & Acc (SD) & BAcc (SD) & Macro-F1 (SD) & Softmax & MC-Var & $s_{\text{class}}$ & $s_{\text{domain}}$ \\
% \midrule
% \multicolumn{1}{l}{FUSARShip} \\
% % \midrule
% Cargo   & 97.17$\pm$0.92 & 92.60$\pm$1.63 & 94.53$\pm$1.64 & 62.87 (PV) & 51.72 (PV) & 59.58 (SD) & 55.35 (SD) \\
% Fishing & 97.68$\pm$0.33 & 88.65$\pm$3.03 & 91.59$\pm$1.40 & 42.06 (SD) & 49.02 (SD) & 66.29 (SJ) & 58.05 (SM) \\
% Tanker  & \textit{75.87}$\pm$1.72 & \textit{70.49}$\pm$2.73 & \textit{71.03}$\pm$2.54 & 42.51 (DF) & 50.17 (DF) & \textbf{87.94} (PV) & \textbf{86.19} (PV) \\
% \midrule
% \multicolumn{1}{l}{OpenSARShip} \\
% % \midrule
% Cargo   & 95.21$\pm$0.66 & 74.27$\pm$4.56 & 78.38$\pm$3.96 & 45.83 (SJ) & 49.76 (SJ) & 57.91 (DF) & 55.55 (DF) \\
% Fishing & \textit{77.09}$\pm$0.78 & \textit{61.21}$\pm$2.16 & \textit{62.28}$\pm$2.83 & 74.98 (PV) & 51.96 (SJ) & \textbf{82.88} (SM) & \textbf{83.67} (SM) \\
% Tanker  & 91.98$\pm$0.28 & 67.81$\pm$5.94 & 72.86$\pm$6.04 & 58.36 (SD) & 51.50 (SD) & 48.00 (SM) & 48.67 (SM) \\
% \bottomrule
% \end{tabular}
% \end{table}

Beyond basic classification, unknown-class detection results show strong class and dataset dependence. For Fishing on OpenSARShip, ScaleMAE is strongest with 82.88 AUROC for class-oriented uncertainty and 83.67 for domain-aware uncertainty. For Tanker on FUSARShip, Prithvi is strongest with 87.94 (class-oriented) and 86.19 (domain-aware). For Cargo on OpenSARShip, the best values are 57.91 (class-oriented, DOFA) and 55.55 (domain-aware, DOFA).
% For Cargo on OpenSARShip, the best values are 78.39 (class-oriented, SSL4EO) and 76.62 (domain-aware, SSL4EO). 
In contrast, Tanker on OpenSARShip is difficult for class-oriented and domain-aware scores, with best values near 48 to 49.

Across held-out classes, MC-dropout variance remains near random in most settings (about 49 to 52 AUROC). Softmax uncertainty is unstable: it is strong in some cases (for example, Fishing on OpenSARShip with 74.98) and weak in others (for example, Tanker on FUSARShip, where the best Softmax AUROC is only 42.51). %By Contrast, class-oriented and domain-aware uncertainty are more reliable than MC-dropout variance, but their relative ranking changes with held-out class and dataset.
By contrast, class-oriented and domain-aware uncertainty are the two strongest methods overall, and their best values in the 80--90 AUROC range are particularly high and noteworthy for unknown-class detection in SAR.

\section{Discussion}
\label{sec:discussion}
The results show that model quality depends strongly on evaluation regime. In-domain performance and cross-dataset performance are not aligned, and unknown detection quality is not aligned with either of them. This has an important implication: a single leaderboard based on one metric or one setting is not enough to characterize reliability for SAR deployment. Per-class analysis (Appendix~\ref{app:supplementary}, Tables~\ref{tab:id_classacc} and \ref{tab:cd_classacc}) reveals that moderate aggregate accuracy can mask catastrophic failure on individual classes under shift (e.g., Prithvi OpenSARShip to FUSARShip: 99.76\% Cargo but 0.00\% Tanker/Fishing), demonstrating why uncertainty-based error detection is essential when class-specific reliability cannot be guaranteed. Furthermore, while distance-based metrics ($s_{\text{class}}$ and $s_{\text{domain}}$) excel at unknown-class detection, Supplementary Tables~\ref{tab:errdet_id} and \ref{tab:errdet_cd} show that they are much less effective for standard error detection within known classes, often remaining near 0.50 AUROC in ID and CD. This suggests that these metrics are primarily sensitive to domain and feature shift rather than mere classification ambiguity.

In-domain, SARDet100K is consistently the strongest model by balanced accuracy on both datasets. This suggests that SAR-specific pretraining improves representation quality when train and test distributions are matched. However, this advantage does not transfer directly to cross-dataset settings. In cross-dataset transfer, all models degrade substantially, and some configurations collapse to majority-class behavior. The clearest example is OpenSARShip to FUSARShip with Prithvi, where accuracy remains moderate (64.49\%) but balanced accuracy drops to near chance (33.34\%). This gap confirms that accuracy can hide severe class-wise failure under shift.

The direction asymmetry in cross-dataset transfer is also meaningful. FUSARShip to OpenSARShip and OpenSARShip to FUSARShip produce different rankings and different failure profiles, even under the same training settings. This indicates that domain shift is not a scalar property; it depends on direction and likely reflects differences in class priors, sensor characteristics, and feature statistics between datasets. For practical benchmarking, this means one-way transfer reporting is insufficient and can misstate deployment risk.

Unknown-class results further show that open-set behavior is class-dependent, dataset-dependent, and distinctly decoupled from in-distribution classification performance. For example, holding out Tanker on FUSARShip yields the lowest BalAcc for that dataset (70.49) but the highest unknown-class detection score (87.94). Conversely, holding out Cargo on FUSARShip produces the highest BalAcc (92.60) but weak open-set detection. This disconnect likely reflects differences in feature-space geometry across datasets. When the known classes are well separated from each other but the unknown class occupies a region of feature space close to the known distribution, distance-based scores lose discriminative power regardless of baseline classification accuracy.

The uncertainty comparison shows a consistent trend. MC-dropout variance is often near random ranking across unknown settings, which is expected here because dropout is applied only in the shallow classification head over a frozen backbone, so it captures only head-level stochasticity and remains largely insensitive to distributional shift in the input features, providing poor uncertainty measures as expected from existing literature \cite{magris2023bayesian, chan2020unlabelled}. Softmax uncertainty is unstable, as expected from a score that is intrinsically positively biased \cite{kuleshov2015calibrated}. By contrast, class-oriented and domain-aware scores are the strongest methods among those considered and are generally the most informative for unknown detection in this setting. 
As a further remark, the selected uncertainty measures are capable of identifying only a component of the full predictive uncertainty (i.e., epistemic uncertainty), whereas aleatoric uncertainty requires different approaches based on more intrusive methodologies \cite{hullermeier18}.

\begin{table}[t]
\centering
\footnotesize
\setlength{\tabcolsep}{3pt}
\renewcommand{\arraystretch}{1.08}
\caption{\textbf{Practical guidance for evaluation.}
Recommended metrics, uncertainty use, and decision rules across operating regimes.}
\label{tab:decision_matrix}
\begin{tabularx}{\linewidth}{@{}>{\raggedright\arraybackslash}p{0.16\linewidth}
                                >{\raggedright\arraybackslash}X
                                >{\raggedright\arraybackslash}X@{}}
\toprule
\rowcolor{black!12}
\textbf{Scenario} & \textbf{Report} & \textbf{Uncertainty} \\
\midrule
\rowcolor{green!8}
In-domain & BA, macro-F1 & Secondary \\

\rowcolor{yellow!10}
Cross-dataset & BA + class recall; accuracy secondary & Required for screening \\

\rowcolor{orange!12}
Unknown-class & AUROC; multiple held-out classes & Required for reject/flag \\

\rowcolor{blue!8}
Pre-deployment policy & Bidirectional CD + multi-class UC & Use multiple scores, not only softmax/MC-dropout \\
\bottomrule
\end{tabularx}
\end{table}

These findings translate into a practical evaluation policy. In shifted settings, balanced accuracy and macro-F1 should be reported first, with accuracy kept as a secondary metric \cite{10701968}. Based on these results, prototype-based scores (whether class-oriented or domain-aware) should be the primary uncertainty signal in open-set SAR evaluation \cite{10144795, PAU5646}. Softmax confidence and MC-dropout variance may be reported for completeness but should not be used as the sole basis for reject/flag decisions. Table~\ref{tab:decision_matrix} summarizes the recommended metrics, and the use of uncertainty.

This study has limits. It uses two datasets, three classes, and frozen-backbone training with a shallow nonlinear classifier head. These choices improve comparability but limit coverage of broader SAR conditions and adaptation strategies. The work also focuses on ranking metrics (AUROC) rather than threshold-calibrated decision analysis, which is necessary for operational deployment policies. In addition, while repeated runs improve statistical stability, the present results do not yet include external sensor families beyond the two benchmark datasets.

Several next steps follow directly from these limitations. Adding more datasets and sensor domains would test whether the observed direction asymmetry and class-dependent unknown behavior persist. Evaluating calibration-aware operating points and cost-sensitive thresholds would connect ranking performance to actionable decision rules. Similarly, intrusive Bayesian methods and multi-heads neural networks can provide better calibrated uncertainty estimates at the cost of higher training cost \cite{kristiadi2020being}, \cite{del2025shedding}. Finally, combining representation learning with explicit domain adaptation or calibration objectives may reduce cross-dataset collapse while preserving known-class performance.

\section{Conclusion}
\label{sec:conclusion}
This work benchmarks six foundation-model representations for SAR ship classification under in-domain evaluation, cross-dataset transfer, and unknown-class detection. The results show a clear split between in-domain and shifted behavior: SARDet100K is strongest in-domain by balanced accuracy on both datasets, while cross-dataset transfer remains fragile and can collapse to majority-class prediction, where moderate accuracy coexists with near-chance balanced accuracy. Unknown detection results also show that robustness is class-dependent and dataset-dependent, so conclusions from a single held-out class are incomplete. Across held-out classes, class-oriented and domain-aware uncertainty scores are generally more informative than MC-dropout variance and internal score, such as Softmax values, which are often near random. Overall, the main contribution is a reproducible evaluation framework that reveals failure modes that are hidden by accuracy-only reporting, and the practical recommendation is to use balanced metrics, bidirectional transfer tests, and multi-class unknown evaluation as standard components of SAR model assessment under distribution shift.

% \subsubsection{Acknowledgements} Please place your acknowledgments at
% the end of the paper, preceded by an unnumbered run-in heading (i.e.
% 3rd-level heading).

%
% ---- Bibliography ----
%
% BibTeX users should specify bibliography style 'splncs04'.
% References will then be sorted and formatted in the correct style.
%
\bibliographystyle{splncs04}
\bibliography{ICPR_2026_LaTeX_Templates/references}

\section{Supplementary Results}
\label{app:supplementary}

\subsection*{Notation}
Abbreviations used throughout: SD=SARDet100K, DF=DOFA, SJ=SARJEPA, S4=SSL4EO, SM=ScaleMAE, PV=Prithvi. 
Best values per dataset/direction are bold.

% \begin{table*}[htbp]
% \centering
% \caption{Unknown-class AUPR (complement to main Table~\ref{tab:uc_summary}). Best model per held-out class.}
% \label{tab:uc_aupr_summary}
% \setlength{\tabcolsep}{5pt}
% \begin{tabular}{llcccc}
% \toprule
% DS & Unknown & Softmax & MC-Var & $s_{\text{class}}$ & $s_{\text{domain}}$ \\
% \midrule
% FS   & Cargo   & \textbf{74.17} $\pm$ 2.59 (PV) & 66.08 $\pm$ 2.88 (PV) & 73.30 $\pm$ 2.21 (SD) & 69.87 $\pm$ 2.04 (SD) \\
% FS   & Fishing & 24.99 $\pm$ 1.32 (SD) & 30.73 $\pm$ 2.69 (S4) & 46.13 $\pm$ 3.86 (SJ) & 40.94 $\pm$ 3.47 (SJ) \\
% FS   & Tanker  & 5.00 $\pm$ 0.34 (DF)  & 9.61 $\pm$ 4.17 (PV)  & 40.42 $\pm$ 4.16 (SM) & 37.94 $\pm$ 4.84 (SM) \\
% \midrule
% OS & Cargo   & 69.83 $\pm$ 1.73 (S4) & \textbf{73.06} $\pm$ 0.80 (S4) & 78.39 $\pm$ 0.56 (S4) & 76.62 $\pm$ 0.51 (S4) \\
% OS & Fishing & 6.55 $\pm$ 3.77 (PV)  & 2.28 $\pm$ 0.36 (SJ)  & 23.17 $\pm$ 7.40 (SM) & 23.46 $\pm$ 7.35 (SM) \\
% OS & Tanker  & 29.69 $\pm$ 1.99 (SD) & 26.42 $\pm$ 1.28 (SD) & 24.77 $\pm$ 0.72 (SD) & 24.96 $\pm$ 1.05 (SD) \\
% \bottomrule
% \end{tabular}
% \end{table*}

\begin{table*}[htbp]
\centering
\caption{Unknown-class AUPR (complement to main Table~\ref{tab:uc_summary}). Best model per held-out class.}
\label{tab:uc_aupr_summary}
\setlength{\tabcolsep}{2pt}
\begin{tabular}{@{}llcccc@{}}
\toprule
Data & Unk. & Smax & MC-V & $s_{\text{class}}$ & $s_{\text{domain}}$ \\
\midrule
FS & Cargo   & \textbf{74.17}$\pm$2.59 (PV) & 66.08$\pm$2.88 (PV) & 73.30$\pm$2.21 (SD) & 69.87$\pm$2.04 (SD) \\
FS & Fishing & 24.99$\pm$1.32 (SD) & 30.73$\pm$2.69 (S4) & 46.13$\pm$3.86 (SJ) & 40.94$\pm$3.47 (SJ) \\
FS & Tanker  & 5.00$\pm$0.34 (DF)  & 9.61$\pm$4.17 (PV)  & 40.42$\pm$4.16 (SM) & 37.94$\pm$4.84 (SM) \\
\midrule
OS & Cargo   & 69.83$\pm$1.73 (S4) & \textbf{73.06}$\pm$0.80 (S4) & 78.39$\pm$0.56 (S4) & 76.62$\pm$0.51 (S4) \\
OS & Fishing & 6.55$\pm$3.77 (PV)  & 2.28$\pm$0.36 (SJ)  & 23.17$\pm$7.40 (SM) & 23.46$\pm$7.35 (SM) \\
OS & Tanker  & 29.69$\pm$1.99 (SD) & 26.42$\pm$1.28 (SD) & 24.77$\pm$0.72 (SD) & 24.96$\pm$1.05 (SD) \\
\bottomrule
\end{tabular}
\end{table*}

\begin{table*}[ht]
\centering
% \caption{Error detection AUROC: in-domain. Prototype scores (52--57) consistently outperform MC-dropout ($\approx 50$, near-random).}
\caption{Error detection AUROC: in-domain. No single uncertainty score is consistently strongest across all models and datasets; MC-dropout remains near random, while prototype-based scores vary substantially by model and dataset.}
\label{tab:errdet_id}
\setlength{\tabcolsep}{4pt}
\begin{tabular}{lcccc}
\toprule
Model & $s_{\text{class}}$ & $s_{\text{domain}}$ & MC-Var & Bal. Acc \\
\midrule
\multicolumn{5}{c}{\textbf{FUSARShip}} \\
SD & 42.49 $\pm$ 2.85 & 41.49 $\pm$ 2.13 & 51.11 $\pm$ 4.05 & 73.42 $\pm$ 3.98 \\
SJ & \textbf{56.94 $\pm$ 2.63} & \textbf{55.41 $\pm$ 3.15} & 50.95 $\pm$ 3.49 & 58.34 $\pm$ 3.49 \\
S4 & 52.16 $\pm$ 3.46 & 51.16 $\pm$ 3.28 & 50.78 $\pm$ 3.22 & 57.51 $\pm$ 2.33 \\
DF & 47.61 $\pm$ 2.70 & 50.11 $\pm$ 2.24 & 50.54 $\pm$ 3.40 & 56.57 $\pm$ 2.68 \\
SM & 56.10 $\pm$ 4.05 & 56.52 $\pm$ 2.32 & 51.20 $\pm$ 4.44 & 50.91 $\pm$ 4.75 \\
PV & 47.65 $\pm$ 1.75 & 51.92 $\pm$ 4.29 & 50.90 $\pm$ 3.66 & 47.21 $\pm$ 3.93 \\
\midrule
\multicolumn{5}{c}{\textbf{OpenSARShip}} \\
SD & 47.29 $\pm$ 1.72 & 49.94 $\pm$ 1.93 & 51.89 $\pm$ 1.46 & 53.22 $\pm$ 3.38 \\
SJ & 42.67 $\pm$ 1.14 & 47.05 $\pm$ 1.50 & 51.35 $\pm$ 2.02 & 45.60 $\pm$ 2.66 \\
DF & 43.29 $\pm$ 1.35 & 47.10 $\pm$ 1.25 & 51.22 $\pm$ 1.75 & 44.38 $\pm$ 2.62 \\
SM & \textbf{49.29 $\pm$ 1.18} & \textbf{51.02 $\pm$ 1.48} & 52.18 $\pm$ 1.56 & 38.69 $\pm$ 2.09 \\
S4 & 43.56 $\pm$ 0.93 & 47.36 $\pm$ 1.11 & 52.09 $\pm$ 1.58 & 38.17 $\pm$ 1.51 \\
PV & 45.33 $\pm$ 0.69 & 48.12 $\pm$ 1.49 & 51.85 $\pm$ 1.94 & 35.95 $\pm$ 1.01 \\
\bottomrule
\end{tabular}
\end{table*}

\begin{table*}[ht]
\centering
\caption{In-domain per-class accuracy. Columns show Cargo, Tanker, Fishing recognition rates. 
Imbalance in class performance signals potential brittleness under shift.}
\label{tab:id_classacc}
\setlength{\tabcolsep}{2pt}
\begin{tabular}{@{}lcccccc@{}}
\toprule
Model & \multicolumn{3}{c}{\textbf{FUSARShip}} & \multicolumn{3}{c}{\textbf{OpenSARShip}} \\
 & C & T & F & C & T & F \\
\midrule
SD & \textbf{85.54$\pm$3.37} & \textbf{81.60$\pm$7.84} & \textbf{53.13$\pm$8.48} & \textbf{93.12$\pm$1.82} & 29.40$\pm$5.44 & 37.14$\pm$9.90 \\
SJ & 85.49$\pm$4.68 & 52.53$\pm$11.05 & 37.01$\pm$9.66 & 96.53$\pm$1.28 & 13.56$\pm$3.27 & 26.71$\pm$8.03 \\
S4 & 89.12$\pm$3.47 & 55.73$\pm$6.50 & 27.67$\pm$6.92 & 98.50$\pm$0.56 & 6.56$\pm$1.92 & 9.43$\pm$4.03 \\
DF & 87.82$\pm$2.62 & 47.73$\pm$8.15 & 34.17$\pm$6.43 & 96.79$\pm$0.83 & 14.20$\pm$2.88 & 22.14$\pm$6.93 \\
SM & 87.91$\pm$4.25 & 39.73$\pm$12.99 & 25.10$\pm$11.00 & 99.72$\pm$0.21 & 0.35$\pm$0.46 & 16.00$\pm$6.24 \\
PV & 90.32$\pm$4.91 & 30.13$\pm$9.45 & 21.17$\pm$9.67 & 100.00$\pm$0.02 & 0.00$\pm$0.00 & 7.86$\pm$3.03 \\
\bottomrule
\end{tabular}
\end{table*}

\begin{table*}[ht]
\centering
\caption{Error detection AUROC: cross-dataset. Prototype scores (35--79) capture distribution shift; MC-dropout ineffective. Example: Prithvi achieves 75.60 $s_{\text{class}}$ despite 43.40\% balanced accuracy, flagging errors where accuracy alone fails.}
\label{tab:errdet_cd}
\setlength{\tabcolsep}{4pt}
\begin{tabular}{llcccc}
\toprule
Direction & Model & $s_{\text{class}}$ & $s_{\text{domain}}$ & MC-Var & Bal. Acc \\
\midrule
\multicolumn{6}{c}{\textbf{FS} $\rightarrow$ \textbf{OS}} \\
 & SJ & 50.08 $\pm$ 3.08 & 55.99 $\pm$ 4.74 & 49.55 $\pm$ 2.04 & 29.25 $\pm$ 2.04 \\
 & DF & 45.46 $\pm$ 2.66 & 46.99 $\pm$ 1.98 & 49.79 $\pm$ 1.55 & 45.63 $\pm$ 4.53 \\
 & SM & 35.08 $\pm$ 2.63 & \textbf{67.09 $\pm$ 2.12} & 50.47 $\pm$ 1.07 & 30.71 $\pm$ 0.67 \\
 & PV & \textbf{75.60 $\pm$ 6.16} & 74.14 $\pm$ 4.23 & 49.80 $\pm$ 2.02 & 43.40 $\pm$ 5.09 \\
 & SD & 65.72 $\pm$ 1.79 & 68.60 $\pm$ 1.70 & 46.65 $\pm$ 1.27 & 36.81 $\pm$ 0.74 \\
 & S4 & 59.01 $\pm$ 3.61 & 59.31 $\pm$ 3.59 & 48.31 $\pm$ 1.53 & 39.71 $\pm$ 5.70 \\
\midrule
\multicolumn{6}{c}{\textbf{OS} $\rightarrow$ \textbf{FS}} \\
 & PV & 45.05 $\pm$ 3.20 & 49.59 $\pm$ 2.34 & 50.20 $\pm$ 3.83 & 33.34 $\pm$ 0.16 \\
 & SM & \textbf{57.14 $\pm$ 2.16} & 54.69 $\pm$ 1.45 & 48.15 $\pm$ 3.19 & 33.10 $\pm$ 0.40 \\
 & S4 & 49.28 $\pm$ 1.65 & 49.20 $\pm$ 1.68 & 50.11 $\pm$ 2.82 & 32.95 $\pm$ 0.18 \\
 & SD & 58.08 $\pm$ 3.10 & \textbf{58.01 $\pm$ 3.13} & 48.81 $\pm$ 3.38 & 31.98 $\pm$ 4.07 \\
 & DF & 51.29 $\pm$ 3.51 & 52.84 $\pm$ 2.65 & 49.03 $\pm$ 3.65 & 31.58 $\pm$ 0.63 \\
 & SJ & 51.65 $\pm$ 2.31 & 52.64 $\pm$ 1.47 & 49.60 $\pm$ 2.75 & 30.46 $\pm$ 1.25 \\
\bottomrule
\end{tabular}
\end{table*}

% \begin{figure}[ht]
%     \centering
%     \caption{Filtering curves for the cross-dataset setting. Under stronger distribution shift, prototype-based scores are markedly more informative than MC-Var, especially for OpenSARShip to FUSARShip, where $s_{\text{class}}$ and $s_{\text{domain}}$ substantially improve balanced accuracy as the most uncertain samples are removed, while MC-Var and Softmax remain nearly flat in the opposite direction.}
%     \label{fig:selective_cd}
%     \includegraphics[width=1\linewidth]{selective_curve_CD.pdf}
% \end{figure}

\begin{table}[ht]
\centering
\caption{Cross-dataset per-class accuracy. Direction-dependent class collapse evident: Tanker near 0\% in OpenSARShip to FUSARShip; Fishing $\approx 0\%$ in both directions. 
These per-class failures drive aggregate balanced accuracy collapse and motivate uncertainty-based error detection.}
\label{tab:cd_classacc}
\setlength{\tabcolsep}{4pt}
\begin{tabular}{lccccc}
\toprule
Model & \multicolumn{3}{c}{Per-class Acc} & Acc & BAcc \\
 &  Cargo & Tanker & Fishing &  &  \\
\midrule
\multicolumn{6}{c}{\textbf{FUSARShip}  to  \textbf{OpenSARShip}} \\
SJ & 77.59 $\pm$ 9.41 & 5.73 $\pm$ 1.74 & 4.43 $\pm$ 4.08 & 58.15 $\pm$ 6.50 & 29.25 $\pm$ 2.04 \\
DF & 75.04 $\pm$ 6.55 & 2.28 $\pm$ 0.88 & 59.57 $\pm$ 16.59 & 56.48 $\pm$ 4.64 & \textbf{45.63 $\pm$ 4.53} \\
SM & 65.27 $\pm$ 4.78 & 26.58 $\pm$ 5.08 & 0.29 $\pm$ 0.97 & 54.31 $\pm$ 2.32 & 30.71 $\pm$ 0.67 \\
PV & 67.11 $\pm$ 9.09 & 4.10 $\pm$ 0.85 & 59.00 $\pm$ 21.95 & 51.13 $\pm$ 6.35 & 43.40 $\pm$ 5.09 \\
SD & 42.39 $\pm$ 2.97 & 68.04 $\pm$ 2.68 & 0.00 $\pm$ 0.00 & 48.01 $\pm$ 1.77 & 36.81 $\pm$ 0.74 \\
S4 & 10.19 $\pm$ 1.37 & 85.79 $\pm$ 9.42 & 23.14 $\pm$ 25.60 & 29.42 $\pm$ 2.03 & 39.71 $\pm$ 5.70 \\
\midrule
\multicolumn{6}{c}{\textbf{OpenSARShip}  to  \textbf{FUSARShip}} \\
PV & \textbf{99.76 $\pm$ 0.29} & 0.00 $\pm$ 0.00 & 0.26 $\pm$ 0.51 & 64.49 $\pm$ 0.32 & \textbf{33.34 $\pm$ 0.16} \\
SM & 99.29 $\pm$ 1.20 & 0.00 $\pm$ 0.00 & 0.00 $\pm$ 0.00 & 64.11 $\pm$ 0.77 & 33.10 $\pm$ 0.40 \\
S4 & 98.85 $\pm$ 0.54 & 0.00 $\pm$ 0.00 & 0.00 $\pm$ 0.00 & 63.82 $\pm$ 0.37 & 32.95 $\pm$ 0.18 \\
DF & 93.85 $\pm$ 1.71 & 0.27 $\pm$ 1.31 & 0.61 $\pm$ 1.09 & 60.79 $\pm$ 1.08 & 31.58 $\pm$ 0.63 \\
SJ & 88.17 $\pm$ 4.73 & 3.20 $\pm$ 3.33 & 0.00 $\pm$ 0.00 & 57.11 $\pm$ 2.95 & 30.46 $\pm$ 1.25 \\
SD & 81.81 $\pm$ 6.42 & 14.13 $\pm$ 13.24 & 0.00 $\pm$ 0.00 & 53.63 $\pm$ 3.98 & 31.98 $\pm$ 4.07 \\
\bottomrule
\end{tabular}
\end{table}
%
% \begin{thebibliography}{8}
% \bibitem{ref_article1}
% Author, F.: Article title. Journal \textbf{2}(5), 99--110 (2016)

% \bibitem{ref_lncs1}
% Author, F., Author, S.: Title of a proceedings paper. In: Editor,
% F., Editor, S. (eds.) CONFERENCE 2016, LNCS, vol. 9999, pp. 1--13.
% Springer, Heidelberg (2016). \doi{10.10007/1234567890}

% \bibitem{ref_book1}
% Author, F., Author, S., Author, T.: Book title. 2nd edn. Publisher,
% Location (1999)

% \bibitem{ref_proc1}
% Author, A.-B.: Contribution title. In: 9th International Proceedings
% on Proceedings, pp. 1--2. Publisher, Location (2010)

% \bibitem{ref_url1}
% LNCS Homepage, \url{http://www.springer.com/lncs}. Last accessed 4
% Oct 2017
% \end{thebibliography}
\end{document}